\documentclass[letterpaper, 10 pt, conference]{ieeeconf}  

\IEEEoverridecommandlockouts                              

\usepackage{graphics} 
\usepackage{epsfig} 
\usepackage{times} 
\usepackage{amsmath} 
\usepackage{amssymb}  
\usepackage{algorithm}  
\usepackage{algorithmic}
\usepackage{booktabs}
\usepackage{xcolor}
\usepackage{multirow}
\usepackage{cite}
\usepackage{xspace}
\usepackage{hyperref}

\newcommand\mypar[1]{\par\vspace{1.0mm}\noindent\textbf{#1}\;\;}

\newcommand\reftab[1]{Table~\ref{#1}}
\newcommand\reffig[1]{Fig.~\ref{#1}}
\newcommand\refsec[1]{Sec.~\ref{#1}}

\DeclareMathOperator*{\argmax}{arg\,max}
\DeclareMathOperator*{\argmin}{arg\,min}
\let\emptyset\varnothing

\newcommand\iou{{\text{IoU}}}
\newcommand\miou{{\text{mIoU}}}
\newcommand\mioubase{{\miou_\base}}
\newcommand\miounovel{{\miou_\novel}}

\newcommand{\finetune}{{\text{GFSS}}}
\newcommand{\finetunefrz}{{\finetune}}
\newcommand{\finetunedyn}{{\finetune_{\text{dyn}}}}
\def\distill{LwF\xspace}
\def\unbiasloss{UBLoss\xspace}
\def\semanticvector{SemVec\xspace}
\def\ourmethod{TeFF\xspace}
\def\ourdataset{SemanticKITTI\xspace}

\newcommand{\lovasz}{{\text{Lov}\acute{\text{a}}\text{sz} }}

\newcommand\base{{base}}
\newcommand\novel{{novel}}

\newcommand\inputspace{\mathcal{X}}

\newcommand\cspace{\mathcal{C}}
\newcommand\cspacebase{\cspace_\base}
\newcommand\cspacenovel{\cspace_\novel}
\newcommand\tspace{\mathcal{T}}
\newcommand\tspacebase{\tspace_\base}
\newcommand\tspacenovel{\tspace_\novel}

\newcommand\realset{\mathbb{R}}

\newcommand\track{\mathbf{Track}}
\newcommand{\param}{{\theta}}
\newcommand\parambase{{\param_\base}}
\newcommand\paramnovel{{\param_\novel}}

\newcommand\model{f_\param}
\newcommand\modelbase{f_\parambase}
\newcommand\modelnovel{f_\paramnovel}

\newcommand{\probability}{p}
\newcommand{\unbiasprobability}{\Tilde{\probability}}

\newcommand\loss{\mathcal{L}}
\newcommand\xentropyloss{\loss_{CE}}
\newcommand\weightxentropyloss{\xentropyloss^w}
\newcommand\unbiasxentropyloss{\Tilde{\loss}_{CE}}

\newcommand\lovaszloss{\loss_{LS}}
\newcommand\distillloss{\loss_{DS}}
\newcommand\unbiasdistillloss{\Tilde{\loss}_{DS}}

\newcommand\jaccard{\mathcal{J}}

\newcommand\inputx{X}
\newcommand\predict{\hat{Y}}
\newcommand\labely{Y}

\newcommand\emphsiz{\textbf}
\newcommand\backbone{\mathbf{BN}}
\newcommand\classhead{\mathbf{CLS}}
\newcommand\classheadbase{\classhead_\base}
\newcommand\classheadnovel{\classhead_\novel}
\newcommand\weight{{W}}

\newcommand{\indexedx}[1]{X^{#1}}
\newcommand{\indexedy}[1]{Y^{#1}}
\newcommand{\indexedpredict}[1]{\predict^{#1}}

\newcommand\rank{{\operatorname{rank}}}

\newcommand\mathif{\text{if }}
\newcommand\otherwise{\text{otherwise }}

\newcommand*\rot{\rotatebox{90}}

\title{\LARGE \bf
TeFF: Tracking-enhanced Forgetting-free Few-shot 3D LiDAR Semantic Segmentation
}

\author{Junbao Zhou$^{1,2}$, Jilin Mei$^{1,\dag}$, Pengze Wu$^{1,2}$, Liang Chen$^{1}$, Fangzhou Zhao$^{1}$, Xijun Zhao$^{3,4}$, Yu Hu$^{1,2,\dag}$
\thanks{*This work was supported by National Natural Science Foundation of China under Grant No.U23B2034, No.62203424, and No.62176250; and in part by the Innovation
Program of Institute of Computing Technology, Chinese Academy of Sciences under Grant
No. 2024000112.}
\thanks{$^{1}$Research Center for Intelligent Computing Systems, Institute of Computing Technology, Chinese Academy of Sciences, Beijing, 100190, China. }%
\thanks{$^{2}$School of Computer Science and Technology, University of Chinese Academy of Sciences, Beijing, 100049, China. }%
\thanks{$^{3}$China North Artificial Intelligence \& Innovation Research Institute.}%
\thanks{$^{4}$Collective Intelligence \& Collaboration Laboratory (CIC). }%
\thanks{$^{\dag}$Correspondence: Jilin Mei, Yu Hu, \{meijilin, huyu\}@ict.ac.cn}%
}

\begin{document}

\maketitle
\thispagestyle{empty}
\pagestyle{empty}

\begin{abstract}

    In autonomous driving, 3D LiDAR plays a crucial role in understanding the vehicle's surroundings. However, the newly emerged, unannotated objects presents few-shot learning problem for semantic segmentation. This paper addresses the limitations of current few-shot semantic segmentation by exploiting the temporal continuity of LiDAR data. Employing a tracking model to generate pseudo-ground-truths from a sequence of LiDAR frames, our method significantly augments the dataset, enhancing the model's ability to learn on novel classes. However, this approach introduces a data imbalance biased to novel data that presents a new challenge of catastrophic forgetting. To mitigate this, we incorporate LoRA, a technique that reduces the number of trainable parameters, thereby preserving the model's performance on base classes while improving its adaptability to novel classes. This work represents a significant step forward in few-shot 3D LiDAR semantic segmentation for autonomous driving. Our code is available at \href{https://github.com/junbao-zhou/Track-no-forgetting}{https://github.com/junbao-zhou/Track-no-forgetting}.

\end{abstract}


\section{INTRODUCTION}

In autonomous driving, 3D LiDAR has been a pivotal sensor due to its proficiency in providing precise 3D position information of surrounding objects \cite{LiMZLCCL21}. This precision is particularly important for semantic segmentation tasks. Semantic segmentation on 3D LiDAR usually leverages deep learning model trained on a large quantity of annotated data \cite{qi2017pointnet, thomas2019kpconv, wu2019pointconv, wang2019dynamic, velivckovic2017graph, qi2017pointnet++, wu2018squeezeseg, wu2019squeezesegv2, milioto2019rangenet++, cortinhal2020salsanext, kong2023rethinking, zhang2020polarnet, aksoy2020salsanet, han2020occuseg, graham20183d, tchapmi2017segcloud, tang2020searching, zhou2020cylinder3d}. However, the autonomous driving scene \cite{behley2019semantickitti} introduces more challenges to deep learning semantic segmentation due to its complexity. In a dynamic environment, the semantic segmentation model may be required to predict newly emerged objects, which is not annotated during training. Additionally, these newly emerged objects (i.e. novel objects) usually lack pixel-level annotations due to the difficulties in collecting and annotating 3D point cloud data. These challenges present the few-shot semantic segmentation problem, which becomes crucial for enhancing the capabilities of autonomous driving systems.

Taking safety into consideration, we extend the few-shot semantic segmentation problem to generalized few-shot semantic segmentation \cite{tian2022generalized, myers2021generalized}. Both settings involve a base training stage with abundant annotated data and a novel data fine-tuning stage with only a few annotated novel classes. However, the generalized one requires the model to be evaluated on both base objects and novel objects while the former one only needs to predict novel objects. Obviously, the generalized few-shot semantic segmentation poses a bigger challenge that needs to be addressed rigorously.

Most of the existing research on generalized few-shot 3D LiDAR semantic segmentation \cite{wu2023generalized} focuses on adapting the model to a few annotated novel data, while preserving the performance on base classes. However, by carefully investigating the LiDAR dataset in autonomous driving scene, we find that the LiDAR data has sequential characteristics from a temporal perspective, which opens a new opportunity for data augmentation. We employ tracking methods \cite{yang2022decoupling} to track LiDAR sequences with a few annotated ground truths, and then append the tracking results to extend the dataset. Those tracking results are considered as ``pseudo ground truths'', which are combined with ground truths to fine-tune the model in the novel stage.

However, a notable problem is raised by the tracking-augmentation that the ``pseudo ground truths'' contain much more points belonging to novel objects than base objects. This data imbalance \cite{johnson2019survey} leads to degradation of accuracy on base classes after novel data fine-tuning, which is known as catastrophic forgetting \cite{kirkpatrick2017overcoming}. To mitigate this issue, we further introduce Low-rank Adaptation (LoRA) \cite{hu2021lora} in few-shot semantic segmentation, which is primarily used in fine-tuning large language models (LLM) \cite{brown2020language, devlin2018bert} on novel tasks. By comparing LLM fine-tuning and novel data fine-tuning in few-shot learning, we find the common characteristics of both tasks is fitting well on novel data while preserving the knowledge of base data. Therefore, by integrating LoRA, our method achieves the goal of \textbf{forgetting-free} and maintains a good balance of accuracy between base classes and novel classes.

Finally, we conduct extensive experiments on \ourdataset \cite{behley2019semantickitti} and show that our method has great improvement on previous few-shot 3D LiDAR semantic segmentation methods and achieves the highest accuracy.

In conclusion, we make the following contributions:
\begin{itemize}
    \item We discover the sequential characteristic of 3D LiDAR data in autonomous driving and leverage it by augmenting the novel data with tracking method \cite{yang2022decoupling}.
    \item We introduce LoRA \cite{hu2021lora} to solve the catastrophic forgetting problem and achieve high accuracy on both base classes and novel classes.
    \item Experiments show that our method outperforms baseline methods and is effective for few-shot 3D LiDAR semantic segmentation, with a noticeable improvement on novel classes.
\end{itemize}

\section{RELATED WORK}

\subsection{3D LiDAR Semantic Segmentation}

3D LiDAR data is unordered and unstructured, which presents a great challenge for segmentation tasks. PointNet \cite{qi2017pointnet} is a milestone in addressing the unstructured problem by proposing a shared MLP network. It extracts features from the whole LiDAR scan directly and then aggregates all the unstructured features through MaxPooling. Unaware of local features, the performance of PointNet \cite{qi2017pointnet} is limited. Following PointNet \cite{qi2017pointnet}, several methods \cite{thomas2019kpconv}, \cite{wu2019pointconv}, \cite{wang2019dynamic}, \cite{velivckovic2017graph} further propose several convolution methods on point cloud to extract local features. Moreover, PointNet++ \cite{qi2017pointnet++} proposes multi-scale sampling rather than convolution to extract multi-scale local features.

As for outdoor scenes, most point cloud segmentation methods transform the unstructured point cloud data into structured 2D data. SqueezeSeg \cite{wu2018squeezeseg},  SqueezeSegv2 \cite{wu2019squeezesegv2}, RangeNet++ \cite{milioto2019rangenet++}, SalsaNext \cite{cortinhal2020salsanext} and
RangeFormer \cite{kong2023rethinking} project the point cloud to a range view (frontal view) image, and utilize 2D convolution network to segment the projected 2D image. Other than range view, bird-eye-view (BEV) is another option to project the point cloud into structured 2D image. PolarNet \cite{zhang2020polarnet}, Salsanet \cite{aksoy2020salsanet} adopt the BEV projection to overcome the data sparsity.

Unlike range-view and BEV, voxelization is another method to convert point cloud into structured data while preserving 3D information. OccuSeg \cite{han2020occuseg}, SSCN \cite{graham20183d}, SegCloud \cite{tchapmi2017segcloud} and SPVConv \cite{tang2020searching} apply 3D convolution networks on voxelized point cloud for LiDAR segmentation. Unlike voxelization, Cylinder3D \cite{zhou2020cylinder3d} proposes cylinder partition of point cloud, which also preserve 3D information.

\subsection{Few-shot 3D LiDAR Semantic Segmentation}

Few-shot semantic segmentation is a problem of making prediction after training on a few labeled novel data. Chen et al. \cite{Chen2020CompositionalPN} proposed a multi-view comparison component that exploits the redundant views of the support set, and extracts prototype features from each view. Zhao et al. \cite{zhao2021few} introduced EdgeConv and self-attention to design a multi-level feature learning network, that learn the geometric and semantic information between points. Lai et al. \cite{Lai2022TacklingBA} explicitly point out that the background ambiguity problem is the main challenge in 3D point cloud semantic segmentation, thus the conventional loss function will lead to degradation of accuracy on few-shot learning.

Previously, we addressed the background ambiguity by introducing unbias cross entropy loss and unbias distillation loss \cite{mei2023few} and we further utilized semantic vectors to enhance the capability of model on fitting novel data \cite{wu2023generalized} .

\subsection{Tracking Methods}

3D Multi-Object Tracking (MOT) involves tracking objects in 3D LiDAR data. Currently, 3D MOT methods can be categorized into ``Tracking-By Detection'' (TBD) and ``Joint Detection and Tracking'' (JDT). Weng et al. \cite{weng20203d}  firstly pioneered the TBD method, tracking by Linear Kalman Filter and 3D IOU, which is simple yet well-performed. SimpleTrack \cite{pang2022simpletrack}, Eagermot \cite{kim2021eagermot} and Camo-mot \cite{wang2023camo} further enhanced the TBD method. Feichtenhofer et al. \cite{feichtenhofer2017detect} firstly proposed JDT method and later, Bergmann et al. \cite{bergmann2019tracking}, Zhang et al. \cite{zhang2020multiple} and  Huang et al. \cite{huang2021joint} improved it. Recently, CenterPoint \cite{yin2021center} proposed a novel tracking method by detecting objects' center and associating them across frames. Currently, TBD methods are generally more precise than JDT methods.

However, 3D MOT methods are not compatible with our task, since in few-shot semantic segmentation, only a few novel objects are annotated and we do not have a well-performed detection model to detect novel objects. Therefore, we adopt video object segmentation (VOS) methods, which don't need semantic information and track each objects by only given annotations in the 1st frame. Early VOS methods \cite{chen2018blazingly, hu2018videomatch, yang2018efficient, yang2020collaborative, voigtlaender2019feelvos} performed feature matching between first frame and following frames, but are challenged by occluded or changing objects. Recently, memory-based VOS method \cite{oh2019video,cheng2022xmem, xie2021efficient, cheng2021rethinking, yang2021associating, yang2022decoupling} have raised research interest and achieved high accuracy on most challenging VOS tasks.

\section{METHODOLOGY}

\subsection{Formulation of Few-shot Learning}
\label{section:formulate_few_shot}
Let $\inputspace$ denotes the input space (i.e. LiDAR data space), each $\inputx \in \inputspace$ denotes a 3D LiDAR scan. Without loss of generality, we assume $|\inputx| = N$, where $N$ is the number of points in $\inputx$. The goal of LiDAR semantic segmentation is to assign a class from class space $\cspace = \{c_1, c_2, ... c_k\}$ to each point in $\inputx$. Therefore, the segmentation result $\labely$ belongs to output space $\cspace^N$. Given a training set $\tspace = \inputspace \times \cspace^N$, the mapping procedure $\inputspace \mapsto \cspace^N$ is performed by a parameterized model $\model$, which  takes $\inputx$ as input and produce point-wise class probability, i.e. $\model : \inputspace \mapsto \realset^{|\cspace| \times N}$. The output mask is further obtained by
\begin{align}
    \labely = \{ \argmax_{c \in \cspace} {\model(c, x_p)} |  p = 1, ... , N \}
\end{align}
where $x_p$ denotes a point in a LiDAR scan with index $p$, and $\model(c, x_p)$ represents the predicted probability of class $c$ at point $x_p$.

In few-shot learning setting, the training process is divided into base stage and novel stage. The class spaces in two training stages are $\cspacebase$ and $\cspacenovel$, respectively, and the two class spaces are disjoint, i.e. $\cspacebase \cap \cspacenovel = \emptyset$. Note that the background or unlabeled class $u$ is excluded in both class spaces. Therefore, the overall class space $\cspace = \{u\} \cup \cspacebase \cup \cspacenovel$, and the training set in base stage and novel stage are $\tspacebase = \inputspace \times \cspacebase^N$ and $\tspacenovel = \inputspace \times \cspacenovel^N$, respectively.

In classical few-shot learning setting, only novel classes $\cspacenovel$ are predicted after two stages of training. However, as for generalized few-shot learning setting, all the classes in $\cspace$ are required to be tested after training. We explicitly point out that generalized few-shot learning setting is more desirable in autonomous driving scene, because with various types of objects on the road, only recognizing novel objects is not sufficient for safety.

We primarily utilize transfer learning \cite{zhuang2020comprehensive} to address few-shot semantic segmentation problem, which can be divided into three steps. In the first step, we train the base model $\modelbase$ on abundant base data through loss function $\loss$:
\begin{align}
    \parambase = \argmin_{\theta} \loss(\inputspace, \cspacebase^N)
\end{align}
Then, we use the base model $\parambase$ to initialize the model in the next step, and fine-tune the model with a few novel data:
\begin{align}
    \paramnovel = \argmin_{\theta} \loss(\inputspace, \cspacenovel^N ; \parambase)
\end{align}

Note that the quantity of labelled data in $\tspacenovel$ is so limited, presenting the \textbf{few-shot} problem. The final step is testing the prediction of all classes with the final model $\model$, where the parameter is simply loaded from the novel model (i.e. $\theta = \paramnovel$). The prediction of each LiDAR scan is obtained by $\predict = \model(\inputx)$.

\begin{figure*}
    \centering
    \includegraphics[width=0.99\textwidth]{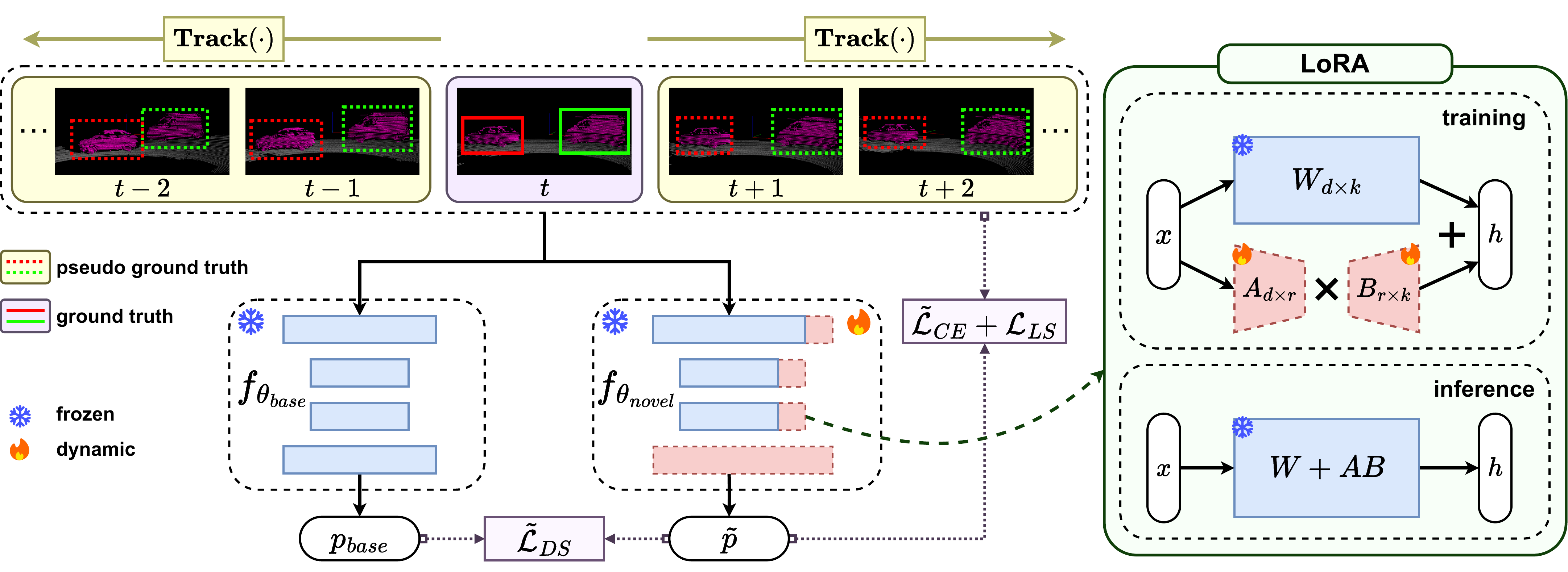}
    \caption{\textbf{Demonstration of our \ourmethod . } In novel data fine-tuning stage, we firstly track each ground truth with tracking model $\track(\cdot)$, forwardly ($t+T$) and backwardly ($t-T$). The tracking results serve as pseudo ground truths and are combined with ground truths to supervise the novel model. We use unbias cross entropy $\unbiasxentropyloss$, unbias distillation $\unbiasdistillloss$ and $\lovasz$ softmax loss $\lovaszloss$ to fine-tune the model. We further apply LoRA to novel model, which reduces the trainable parameters, thus achieving the goal of \textbf{forgetting-free}. }
    \label{fig:method}
\end{figure*}

\subsection{Base Model Training}

In base training stage, we utilize the whole training set of \ourdataset \cite{behley2019semantickitti}, which contains abundant data annotated with classes $c \in \{u\} \cup \cspacebase$. Similar to previous semantic segmentation methods, we use weighted cross entropy loss and the $\lovasz$ softmax loss.

\mypar{Weighted Cross Entropy Loss.} \ourdataset is highly imbalanced annotated, for example, the points of class \emph{road} significantly outnumber the points of other classes. Similar to previous segmentation work \cite{cortinhal2020salsanext}, we incorporate weighted cross entropy loss to overcome this biased distribution. With an input LiDAR scan $\inputx \in \inputspace$ and its corresponding ground truth label $\labely \in \cspace^N$, the conventional cross entropy loss at point $x_p$ is calculated by:
\begin{align}
    \xentropyloss (x_p, y_p) =
    - \log \probability(y_p, x_p)
    \label{eq:xentropy}
\end{align}
where $\probability(y_p, x_p) = \model(y_p, x_p)$ is the predicted probability of the ground truth class $y_p$ at point $x_p$. The weighted cross entropy loss $\weightxentropyloss$ is formulated by:
\begin{align}
    \weightxentropyloss & = \frac{w_{y_p}}{\sum_{c \in \cspace} w_c} \xentropyloss (x_p, y_p) \\
    w_c                 & = \frac{1}{\sqrt{M_c}}
\end{align}
where $M_c$ denotes the number of points belongs to class $c$ in the whole training set.

\mypar{$\lovasz$ Softmax loss.} Similar to previous segmentation work \cite{cortinhal2020salsanext}, we also utilize  $\lovasz$ softmax loss \cite{berman2018lovasz} to maximize the $\miou$ of our model. $\lovasz$ softmax loss is defined as:
\begin{align}
    \lovaszloss  = & \frac{1}{|\cspace|} \sum_{c \in \cspace} \overline{\Delta_{\jaccard_c}}(m(c)), \\
    m(c) =         & \begin{cases}
                         1 - \model(c, x_p) \  & \mathif c = y_p \\
                         \model(c, x_p) \      & \otherwise
                     \end{cases}
\end{align}
where $\jaccard_c$ defines the Jaccard index, and $\overline{\Delta_{\jaccard_c}}$ is the $\lovasz$ extension of the Jaccard index.

The final loss function of the base training stage is :
\begin{align}
    \loss_{base} = \xentropyloss + \lovaszloss
\end{align}

\subsection{Extending Novel Data with Tracking Method}

In autonomous driving scene, the LiDAR data is collected over a continuous time period. Therefore, the LiDAR data is sequential from a temporal perspective. This feature of LiDAR data provides an opportunity for data augmentation via tracking method.

Taking the temporal continuity into consideration, we redefine the dataset as $\tspace = \inputspace \times \cspace^N = \{(\indexedx{t}, \indexedy{t}) | t = 1, 2, ... , T\}$ , where $t$ denotes the timestamp of a LiDAR frame. A tracking model \cite{yang2022decoupling} $\track(\cdot)$ firstly takes an annotated frame $(\indexedx{t}, \indexedy{t})$ as input and extracts its features as $F^t$. Then, $\track(\cdot)$ subsequently takes in the following frames $\{\indexedx{t+1}, \indexedx{t+2}, ... \}$ and produces the segmentation $\{\indexedpredict{t+1}, \indexedpredict{t+2}, ... \}$. This procedure can be defined as:
\begin{align}
    \indexedpredict{t+s}
     & = \track(\indexedx{t+s} | F^t)                        \\
     & = \track(\indexedx{t+s} | \indexedx{t}, \indexedy{t})
\end{align}

Because the temporal continuity still holds in the reverse manner, the tracking model can also predict segmentation in a reverse LiDAR sequence. Therefore, we can obtain the segmentation $\{\indexedpredict{t-1}, \indexedpredict{t-2}, ... \}$ of a reverse LiDAR sequence:
\begin{align}
    \indexedpredict{t-s} = \track(\indexedx{t-s} | \indexedx{t}, \indexedy{t})
\end{align}

We hereby define the segmentation produced by the tracking model as \emphsiz{pseudo ground truth}. With the labelled ground truth $(\indexedx{t}, \indexedy{t})$, we combine them and construct a augmented dataset $\hat{\tspace} = \inputspace \times \cspace^N$, where:
\begin{align}
    \hat{\tspace} = \{(\indexedx{t}, \indexedy{t}), (\indexedx{t+s}, \indexedpredict{t+s}) | s = -T, ..., -1, 1, ... , T\}
\end{align}
and $T$ denotes the max tracking number of frames.

Since annotation of novel classes is limited in few-shot learning, data augmentation is crucial to prevent over-fitting. The augmented dataset $\hat{\tspace}$ provides more information of novel data and improves the performance in the novel fine-tuning stage.

\subsection{Novel Data fine-tuning}

As is described in \refsec{section:formulate_few_shot}, model $\model$ performs a mapping from input space to point-wise probability, i.e. $\model : \inputspace \mapsto \realset^{|\cspace| \times N}$. This prediction process is accomplished by the combination of a backbone network $\backbone(\cdot)$ and a classification head $\classhead(\cdot)$. The prediction process of model $\model$ can be defined as:
\begin{align}
    \indexedpredict{t}
     & = \model(\indexedx{t})                \\
     & = \classhead(\backbone(\indexedx{t}))
\end{align}

We adopt transfer learning \cite{zhuang2020comprehensive} in few-shot semantic segmentation. In novel fine-tuning stage, we instantiate a new classification head $\classheadnovel(\cdot)$ to predict the novel classes. The output of $\classheadnovel(\cdot)$ is concatenated with the output of the base classification head $\classheadbase(\cdot)$. The parameters in $\backbone(\cdot)$ and $\classheadbase(\cdot)$ are directly loaded from the base model.

\mypar{Mitigating Forgetting with Low Rank Adaptation.} The tracking-augmentation method is able to augment the data with limited annotation extensively. However, the tracking method predominantly focuses on novel classes, which results the ratio of novel classes points in the augmented data significantly higher that in the overall dataset. This imbalance biases the distribution of the augmented dataset and directs the model's learning focus towards novel classes, leading to catastrophic forgetting. This phenomenon is where the model's ability to recognize base classes deteriorates as it increasingly focuses on novel classes. This consequence is particularly problematic as our goal is the generalized few-shot learning problem (outlined in \refsec{section:formulate_few_shot}). We aim to develop a model that maintains high accuracy across both base and novel classes. To mitigate the forgetting issue, we incorporate the Low Rank Adaptation (LoRA) \cite{hu2021lora} approach during the fine-tuning phase.

LoRA is a technique primarily used in large pre-trained models fine-tuning. In traditional fine-tuning, all of the parameters of a pre-trained model are updated during the training process on a new task, which is computationally expensive and time-consuming. Besides, updating the whole model probably leads to catastrophic forgetting, which harms the accuracy on original task after fine-tuning. The core idea of LoRA is to adapt a pre-trained model to a new task with minimal changes, enhancing the model's accuracy on new task while preserving the model's performance on original task.  LoRA achieve this goal by introducing small, trainable weights rather than updating the model's original weights directly. To be more specific, most of the weights can be written in form of matrices, with the denotation $\weight \in \realset^{d \times k}$. During fine-tuning, LoRA constrains the update by decomposition: $\weight+\Delta \weight = \weight + BA$,
where $\rank(A) = \rank(B) \ll \min(d, k)$ . In other words, the trainable parameters in $A$ and $B$ are far less than that in $\weight$. During fine-tuning, $\weight$ is frozen and only $A$ and $B$ receive gradient updates. As for forward pass, the original output is added on the output of $BA$, i.e
\begin{align}
    h = (\weight + \Delta \weight) x = \weight x + BA x
\end{align}
This process ensures that the model's original capabilities are retained while it learns to recognize new classes.

Although not identical, few-shot learning is similar to the large model fine-tuning. As is shown in \reffig{fig:method}, we incorporate LoRA in the novel training stage and significantly reduce the number of trainable parameters. LoRA is only applied to $\backbone(\cdot)$ while $\classheadbase(\cdot)$ and $\classheadnovel(\cdot)$ are kept dynamic. This setting effectively counteract the imbalance issue and mitigate the risk of catastrophic forgetting. This approach not only preserves the model's performance on base tasks but also enhances its accuracy on novel tasks, aligning with our goal of generalized few-shot learning.

\mypar{Unbias Cross Entropy Loss.} Following our previous work \cite{mei2023few, wu2023generalized}, we empirically choose unbias cross entropy loss to mitigate the gap between base training and novel data fine-tuning. The unbias cross entropy loss $\unbiasxentropyloss$ is defined as follows:
\begin{align}
    \unbiasxentropyloss = - \log \unbiasprobability(y_p, x_p)
\end{align}
where
\begin{equation}
    \unbiasprobability(c, x_p) =
    \begin{cases}
        \model(c, x_p)                                  & \mathif c \in \cspacenovel \\
        \sum_{c \in \{u\}\cup\cspacebase}\model(c, x_p) & \otherwise
    \end{cases}
    \label{eq:unbiasxentropy}
\end{equation}


\mypar{Unbias Distillation Loss.} In transfer learning setting, the base model $\modelbase$ serves as a teacher model and supervise the student model (i.e. the novel model $\modelnovel$) through distillation loss $\distillloss$. However, the traditional distillation loss does not take into account that novel objects are annotated as background in the base training stage. Similar to previous few-shot semantic segmentation work \cite{mei2023few, wu2023generalized}, we bridge this gap by using unbias distillation loss $\unbiasdistillloss$, which is defined as:
\begin{align}
    \unbiasdistillloss = - \probability_{base}(y_p, x_p) \log \unbiasprobability(y_p, x_p)
\end{align}
where
\begin{equation}
    \unbiasprobability(c, x_p) =
    \begin{cases}
        \model(c, x_p)                                   & \mathif c \in \cspacebase \\
        \sum_{c \in \{u\}\cup\cspacenovel}\model(c, x_p) & \otherwise
    \end{cases}
    \label{eq:unbiasdistill}
\end{equation}


\section{EXPERIMENTS}

\begin{table}[t]
\centering
\caption{\textbf{Comparison with baselines on SemanticKITTI validation set.}}
\resizebox{0.85\linewidth}{!}{
\begin{tabular}{c|l|ccc}
\toprule
\multirow{2}{*}{Shot}
& Method & $\miou$ & $\mioubase$ & $\miounovel$ \\
\cmidrule{2-5}
& Base Model & - & 58.7 & - \\
\midrule
\multirow{6}{*}{10}
& $\finetunefrz$ & 49.1 & 56.8 & 20.3 \\
& $\finetunedyn$ & 47.8 & 53.5 & 26.3 \\
& \distill & 48.0 & 53.3 & 28.4 \\
& \unbiasloss & 50.1 & 55.7 & 28.8 \\
& \semanticvector & 51.5 & 56.1 & 34.3 \\
& \textbf{\ourmethod (Ours)} & \textbf{55.3} & \textbf{58.7} & \textbf{42.6} \\
\midrule
\multirow{6}{*}{5}
& $\finetunefrz$ & 49.5 & 56.3 & 23.9 \\
& $\finetunedyn$ & 48.1 & 53.6 & 27.5 \\
& \distill & 46.7 & 53.0 & 23.1 \\
& \unbiasloss & 49.6 & 56.4 & 23.8 \\
& \semanticvector & 49.3 & 55.0 & 27.6 \\
& \textbf{\ourmethod (Ours)} & \textbf{53.9} & \textbf{58.1} & \textbf{37.3} \\
\midrule
\multirow{6}{*}{2}
& $\finetunefrz$ & 48.3 & 55.2 & 22.4 \\
& $\finetunedyn$ & 46.4 & 52.5 & 23.5 \\
& \distill & 46.8 & 52.3 & 26.1 \\
& \unbiasloss & 48.9 & 54.8 & 26.6 \\
& \semanticvector & 46.8 & 51.8 & 27.9 \\
& \textbf{\ourmethod (Ours)} & \textbf{53.2} & \textbf{58.6} & \textbf{32.8} \\
\midrule
\multirow{6}{*}{1}
& $\finetunefrz$ & 48.6 & 55.5 & 22.6 \\
& $\finetunedyn$ & 43.9 & 48.2 & 27.9 \\
& \distill & 43.1 & 47.2 & 27.6 \\
& \unbiasloss & 48.5 & 53.8 & 28.8 \\
& \semanticvector & 40.5 & 44.2 & 26.5 \\
& \textbf{\ourmethod (Ours)} & \textbf{52.4} & \textbf{58.0} & \textbf{31.4} \\
\bottomrule
\end{tabular}
}
\label{tab:main_result_val}
\end{table}

\subsection{Dataset and Evaluation Metrics}

\mypar{\ourdataset.} We primarily choose \ourdataset to demonstrate the effectiveness of our method. \ourdataset is a large scale LiDAR dataset features with over 43K 3D LiDAR scan, which are collected in driving scene and provided in sequences. In the semantic segmentation task, \ourdataset provides 20 annotated classes.

Identical to the official config of \ourdataset, we split the dataset into 3 subsets: sequences 00 - 07 and 09 - 10 are used for training, sequences 08 is for validation and sequences 11 - 21 are used for testing. As for few-shot learning setting, we set the \textbf{car}, \textbf{person}, \textbf{bicyclist}, and \textbf{motorcyclist} as the novel classes and the other 16 classes as base classes.

\mypar{Evaluation Metrics.} Similar to our previous work \cite{mei2023few, wu2023generalized}, we evaluate our method with $\miou$, and we further calculate $\mioubase$ and $\miounovel$ for base classes and novel classes separately.

\subsection{Baseline and Implementation Details}

\mypar{Baselines.} We compare our method against several few-shot semantic segmentation methods used in 3D LiDAR data:

\begin{table*}[ht]
\tabcolsep=0.1cm
\centering
\caption{\textbf{Comparison with baselines on SemanticKITTI testing set.}}
\resizebox{0.98\linewidth}{!}{
\begin{tabular}{c|l|ccccccccccccccc|cccc|ccc}
\toprule
{Shot}
& Method & \rot{bicycle} & \rot{motorcycle} & \rot{truck} & \rot{other-vehicle} & \rot{road} & \rot{parking} & \rot{sidewalk} & \rot{other-ground} & \rot{building} & \rot{fence} & \rot{vegetation} & \rot{trunk} & \rot{terrain} & \rot{pole} & \rot{traffic-sign} & \rot{car} & \rot{person} & \rot{bicyclist} & \rot{motorcyclist} & \rot{$\mioubase$} & \rot{$\miounovel$} & \rot{$\miou$} \\
\midrule
\multirow{6}{*}{10}
& $\finetunefrz$ & 30.4 & 26.1 & 27.3 & 21.7 & 90.1 & 57.1 & 73.5 & 27.2 & 84.9 & 53.2 & 77.6 & 60.5 & 63.0 & 49.7 & 55.2 & 80.4 & 0.0 & 1.3 & 0.0 & 53.2 & 20.4 & 46.3 \\
& $\finetunedyn$ & 16.9 & 23.9 & 29.8 & 18.2 & 89.5 & 55.8 & 72.3 & 26.8 & 85.7 & 53.0 & 77.0 & 60.0 & 62.7 & 47.9 & 52.4 & 77.6 & 12.8 & 11.3 & 5.7 & 51.5 & 26.9 & 46.3 \\
& \distill & 19.4 & 24.3 & 32.9 & 18.9 & 89.3 & 54.0 & 70.9 & 24.8 & 85.8 & 52.7 & 77.0 & 59.4 & 61.7 & 45.4 & 50.7 & 78.0 & 13.9 & 12.5 & 5.6 & 51.1 & 27.5 & 46.2 \\
& \unbiasloss & 11.0 & 24.5 & 28.1 & 12.4 & 90.3 & 57.7 & 72.5 & 24.1 & 86.0 & 55.0 & 78.2 & 60.9 & 64.1 & 52.8 & 49.9 & 88.7 & 13.8 & 10.8 & 3.7 & 51.2 & 29.3 & 46.6 \\
& \semanticvector & 33.7 & 25.3 & 26.5 & 21.2 & 90.2 & 57.4 & 72.2 & 27.0 & 84.1 & 50.7 & 76.4 & 61.1 & 63.8 & 49.8 & 48.5 & 87.2 & 21.3 & 11.6 & 3.7 & 52.5 & 31.0 & 48.0 \\
& \textbf{\ourmethod(Ours)} & 19.9 & 29.9 & 26.5 & 20.6 & 90.2 & 59.2 & 72.9 & 28.3 & 85.6 & 55.1 & 79.0 & 62.4 & 64.3 & 53.0 & 56.8 & 89.5 & 24.8 & 18.8 & 6.6 & \textbf{53.6} & \textbf{34.9} & \textbf{49.7} \\
\midrule
\multirow{6}{*}{5}
& $\finetunefrz$ & 27.8 & 30.3 & 25.8 & 23.1 & 90.1 & 56.6 & 72.9 & 26.9 & 85.8 & 53.6 & 77.3 & 59.2 & 62.9 & 45.9 & 56.4 & 87.0 & 4.9 & 0.0 & 0.0 & 53.0 & 23.0 & 46.7 \\
& $\finetunedyn$ & 38.1 & 28.8 & 14.6 & 18.4 & 89.5 & 57.0 & 69.9 & 25.9 & 85.5 & 54.6 & 77.1 & 59.3 & 61.0 & 46.2 & 52.1 & 86.4 & 15.2 & 1.9 & 1.4 & 51.9 & 26.2 & 46.5 \\
& \distill & 26.9 & 15.4 & 4.1 & 14.7 & 88.6 & 56.2 & 69.5 & 26.7 & 85.5 & 54.7 & 75.4 & 58.1 & 59.8 & 46.8 & 52.1 & 85.1 & 11.7 & 1.9 & 1.3 & 49.0 & 25.0 & 43.9 \\
& \unbiasloss & 23.8 & 28.2 & 22.6 & 18.0 & 89.8 & 57.5 & 71.3 & 26.2 & 84.9 & 55.1 & 77.8 & 62.0 & 62.2 & 52.2 & 51.9 & 87.0 & 6.4 & 0.0 & 0.7 & 52.2 & 23.5 & 46.2 \\
& \semanticvector & 36.4 & 28.1 & 24.2 & 23.3 & 89.5 & 53.1 & 70.5 & 27.6 & 84.2 & 48.3 & 76.0 & 61.1 & 62.6 & 38.2 & 54.7 & 88.2 & 11.4 & 2.4 & 1.9 & 51.9 & 26.0 & 46.4 \\
& \textbf{\ourmethod(Ours)} & 31.3 & 30.4 & 26.3 & 20.8 & 90.5 & 60.3 & 72.3 & 27.4 & 85.8 & 55.3 & 78.8 & 62.4 & 62.8 & 53 & 57 & 89.3 & 24.6 & 10.6 & 6.4 & \textbf{54.3} & \textbf{32.7} & \textbf{49.8} \\
\midrule
\multirow{6}{*}{2}
& $\finetunefrz$ & 27.4 & 24.0 & 28.6 & 21.8 & 90.3 & 57.0 & 72.6 & 24.5 & 85.9 & 53.0 & 77.3 & 56.7 & 61.8 & 48.0 & 56.5 & 87.0 & 0.0 & 2.4 & 0.0 & 52.4 & 22.4 & 46.0 \\
& $\finetunedyn$ & 20.5 & 20.9 & 24.8 & 12.5 & 89.1 & 53.5 & 70.7 & 21.2 & 85.3 & 53.0 & 79.4 & 57.7 & 63.8 & 48.6 & 51.9 & 85.8 & 0.8 & 6.1 & 0.5 & 50.2 & 23.3 & 44.5 \\
& \distill & 16.8 & 26.1 & 22.2 & 12.5 & 88.2 & 51.1 & 70.7 & 20.9 & 86.1 & 54.7 & 79.1 & 56.8 & 63.8 & 49.5 & 51.6 & 84.3 & 0.9 & 7.6 & 0.5 & 50.0 & 23.3 & 44.4 \\
& \unbiasloss & 19.4 & 25.9 & 27.9 & 15.5 & 89.9 & 57.5 & 71.8 & 20.7 & 85.6 & 53.9 & 79.4 & 61.8 & 62.8 & 52.6 & 51.3 & 87.0 & 0.3 & 6.5 & 0.6 & 51.7 & 23.6 & 45.8 \\
& \semanticvector & 30.2 & 20.7 & 18.7 & 12.5 & 89.2 & 55.5 & 70.2 & 24.2 & 81.9 & 46.6 & 75.8 & 59.6 & 61.5 & 36.4 & 55.5 & 88.0 & 3.0 & 10.4 & 0.4 & 49.2 & 25.5 & 44.2 \\
& \textbf{\ourmethod(Ours)} & 26.3 & 29 & 28.1 & 21.9 & 90.1 & 60.5 & 72.1 & 26.7 & 85.5 & 54.8 & 79.1 & 62.7 & 63.2 & 53.2 & 54.8 & 88.8 & 14.2 & 11.8 & 6.1 & \textbf{53.9} & \textbf{30.2} & \textbf{48.9} \\
\midrule
\multirow{6}{*}{1}
& $\finetunefrz$ & 15.2 & 17.5 & 29.6 & 19.7 & 89.6 & 56.2 & 71.3 & 10.7 & 82.8 & 47 & 73.2 & 54.6 & 59.5 & 46.4 & 55.4 & 86.1 & 0 & 2.7 & 0 & 48.6 & 22.2 & 43.0 \\
& $\finetunedyn$ & 0 & 17 & 17.8 & 12.4 & 89 & 52.1 & 71.3 & 12.9 & 84.5 & 47.8 & 74.2 & 51.8 & 62.4 & 43.4 & 39.9 & 85.4 & 0.7 & 11 & 0 & 45.1 & 24.3 & 40.7 \\
& \distill & 0 & 19.8 & 21.1 & 12.4 & 89 & 48.6 & 69.2 & 16 & 84.4 & 46.2 & 75.1 & 51.1 & 64.1 & 45 & 48.1 & 85.8 & 0.6 & 12 & 0.1 & 46.0 & 24.6 & 41.5 \\
& \unbiasloss & 3.5 & 25.5 & 26.9 & 17.2 & 89.3 & 54.2 & 70.3 & 8.4 & 83.3 & 50.7 & 76.9 & 55 & 63.9 & 49.2 & 45.2 & 85.2 & 0.4 & 12.3 & 0.2 & 48.0 & 24.5 & 43.0 \\
& \semanticvector & 6.1 & 15.2 & 3.2 & 21.2 & 88.6 & 54.1 & 69.8 & 12.8 & 80.8 & 40.2 & 71.9 & 59.2 & 62.2 & 31.1 & 51.4 & 85.2 & 3.3 & 7.3 & 0.0 & 44.5 & 24.0 & 40.2 \\
& \textbf{\ourmethod(Ours)} & 32.4 & 27.1 & 29 & 21.5 & 89.9 & 58.9 & 71.6 & 18.2 & 85.1 & 52.5 & 77 & 61.4 & 61.2 & 53.1 & 54.1 & 88.4 & 9.7 & 8.7 & 0 & \textbf{52.9} & \textbf{26.7} & \textbf{47.4} \\
\bottomrule
\end{tabular}
}
\label{tab:main_result_test}
\end{table*}

\begin{itemize}
    \item \textbf{$\finetunefrz$.} \cite{myers2021generalized} Generalized few-shot semantic segmentation. After the base training stage, all the parameters in model's backbone are frozen  and only the parameters in classification head receive gradient update.
    \item \textbf{$\finetunedyn$.} Share the same config with GFSS but during novel training stage, the parameters in the backbone are not frozen (i.e. dynamic) and also receive gradient update .
    \item \textbf{\distill.} \cite{li2017learning} Learning without forgetting. During the novel fine-tuning stage, the predicted probabilities of base model are used to supervise the novel model through distillation loss.
    \item \textbf{\unbiasloss.} \cite{mei2023few} Unbias cross entropy and distillation loss. It incorporate the background information and better mitigate the catastrophic forgetting problem.
    \item  \textbf{\semanticvector.} \cite{wu2023generalized} Integrating semantic vectors into few-shot semantic segmentation. During novel fine-tuning stage, semantic vectors are multiplied with the probabilities produced by classifiers, thus incorporating semantic information and enhancing the performance of few-shot learning.
\end{itemize}

\mypar{Model Settings.} We evaluate our method with SalsaNext\cite{cortinhal2020salsanext}, a 3D LiDAR segmentation network, as it is fast and still holds high accuracy on \ourdataset. As for tracking, most of the mainstream 3D LiDAR tracking methods require detection model \cite{li2023poly}. This is not compatible with our few-shot setting as we don't have a model to predict on novel objects until novel fine-tuning stage. Therefore, we adopt a video tracking method, DeAOT\cite{yang2022decoupling}, which does not require semantic information of novel classes and only needs the annotation of objects in the 1st frame. Note that DeAOT model requires 2D images as input, and to make it compatible with 3D LiDAR data, we project the 3D LiDAR into 2D range-view, with resolution $2048 \times 64$. The projected range-view frames are fed subsequently into DeAOT and produce tracking results. The predicted tracking results are reverse-projected back to 3D LiDAR and serve as \textbf{pseudo ground truth}, which will be used in finetuning the SalsaNext model.

\mypar{Training Details.} As is described in \refsec{section:formulate_few_shot}, we adopt transfer learning, which contains two stages: base training and novel data fine-tuning. The base training stage utilize the whole training split while the novel classes (\textbf{car}, \textbf{person}, \textbf{bicyclist}, and \textbf{motorcyclist}) are labeled as background. During novel data fine-tuning, to align with few-shot learning setting, we randomly sample $m$ scans for each novel class (i.e. $m$-shot) from the training split. Notably, considering our method use tracking model to extend data, we particularly ensure a minimum gap of 250 between each LiDAR scan, to avoid data redundancy. On both two training stages, we train the model for 160 epochs with batch size 14, which is sufficient for model to fully fit on the data.

Our proposed method \ourmethod and all the baselines share the same base training stage and start the novel fine-tuning stage with the same base model. By adopting such setting, we ensure all the differences between each method are attributed solely to different novel fine-tuning strategies.

\mypar{\ourmethod Details.} We track each ground truth for 20 frames  with tracking gap 15 (discussed in \refsec{sec:ablation}). As for LoRA, we apply LoRA on all the up-sample blocks and half of the ResNet blocks\cite{cortinhal2020salsanext}, while keep other layers frozen. The $\rank$ in LoRA is set to $1/4$ of the hidden dimension for each layer.

\subsection{Quantitative Analysis}

In \reftab{tab:main_result_val}, we compare our method \ourmethod with previous few-shot semantic segmentation methods in 4 different settings, $\text{shot}=1,2,5$ and $10$. Our method achieves the highest score in all the 4 settings and establishes a new state-of-the-art in few-shot 3D LiDAR semantic segmentation. Notably, \ourmethod not only excels in adapting to novel classes, but also preserves a high score on base classes, effectively addressing the problem of catastrophic forgetting. This capability is especially important in generalized few-shot semantic segmentation for autonomous driving, where all the classes should be accurately predicted due to safety concerns. Our method, \ourmethod, leverages a tracking model to provide sufficient novel data for fine-tuning, and minimizes the catastrophic forgetting by introducing LoRA,  which significantly reduces the trainable parameters.

\reftab{tab:main_result_test} shows the $\iou$ of all the classes on \ourdataset testing split. Our method also performs best on $\mioubase$, $\miounovel$ and $\miou$. Besides, our method also achieves the highest score in most of the classes.

\subsection{Ablation Study}
\label{sec:ablation}

\begin{table}[ht]
\centering
\caption{\textbf{Ablation study of lora config.}}
\resizebox{0.85\linewidth}{!}{
\begin{tabular}{l|ccc}
\toprule
Method & $\miou$ & $\mioubase$ & $\miounovel$ \\
\midrule
Freezing & 50.9 & 58.0 & 24.3 \\
Dynamic & 52.1 & 55.2 & \textbf{40.6} \\
\textbf{LoRA (Ours)} & \textbf{53.2} & \textbf{58.6} & 32.8 \\
\bottomrule
\end{tabular}
}
\label{tab:ablation_lora}
\end{table}

\mypar{Effectiveness of LoRA.} We compare LoRA with two fine-tuning strategies in \reftab{tab:ablation_lora}: (1) Freezing, except the classification head, all the parameters do not receive updates. (2) Dynamic, which tunes all the parameters of the model. LoRA reduces the trainable parameters while not freezes the whole model, thereby preserving good performance on base classes while also fitting well on novel data. Although it is outperformed by the Dynamic strategy on $\miounovel$, it excels in maintaining good balance between base and novel classes, achieving the highest overall $\miou$.

\begin{figure}
    \centering
    \includegraphics[width=0.87\linewidth]{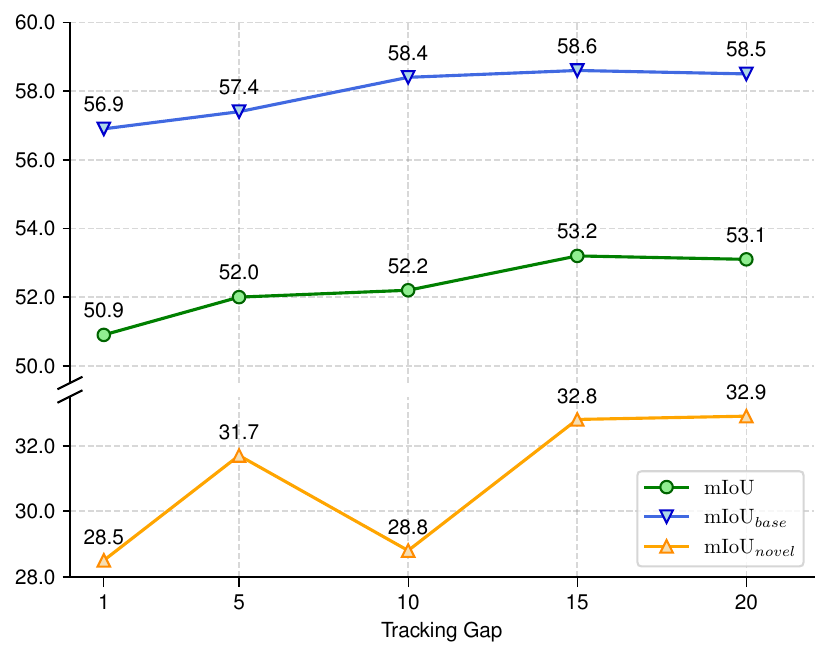}
    \caption{\textbf{Ablation study of tracking gap on \ourdataset validation set.} It is tested with $\text{shot}=2$ and the tracking frame number is 20 (10 forward and 10 backward).}
    \label{fig:ablation_track_gap}
\end{figure}

\mypar{Analysis on the Gap between Tracked Scans.} Although tracking method can provide sufficient novel data, it is not optimal to utilize every tracking result in fine-tuning. Firstly, the adjacent pseudo ground truths are similar, presenting a data redundancy problem, which probably leads to overfitting. Secondly, using too many samples in fine-tuning is computationally expensive and significantly increases the training time. Therefore, we introduce tracking gap, which means selecting a sample every certain scan in a tracking-generated sequence. However, if the tracking-generated sequence goes too long, the quality of tracking results tends to degrade. It means that there is a trade-off in tracking gap preventing it from being unlimitedly large. As shown in \reffig{fig:ablation_track_gap}, the optimal tracking gap is 15, which performs best in the overall $\miou$.

\begin{figure}
    \centering
    \includegraphics[width=0.87\linewidth]{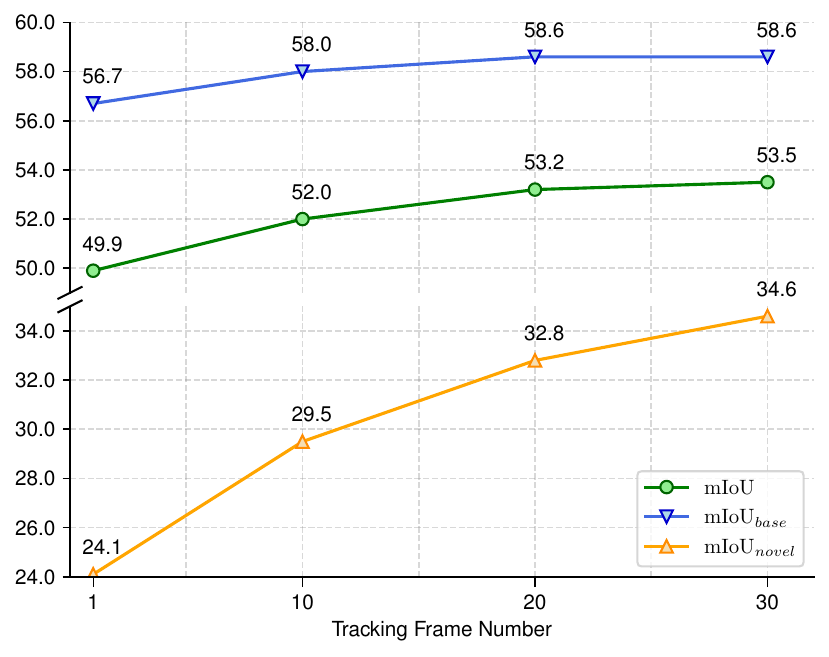}
    \caption{\textbf{Ablation study of tracking frame numbers on \ourdataset validation set.} It is tested with $\text{shot}=2$ and the tracking gap is 15.}
    \label{fig:ablation_track_frame_num}
\end{figure}

\mypar{Analysis on the Number of Tracked Scans.} As shown in \reffig{fig:ablation_track_frame_num}, an increasing number of scans generally improves the overall $\miou$. However, when the tracking frame number exceeds 20, the improvement tends to be minor ($\miou: 53.2 \rightarrow 53.5$). Considering that the tracking model requires much more GPU memory and becomes slow with more tracking frames, we set this value to 20 (10 forward and 10 backward), which is sufficient to demonstrate the effectiveness of our method.

\section{CONCLUSIONS}

In this work, we address the few-shot 3D LiDAR semantic segmentation problem. By exploiting the sequential characteristic of 3D LiDAR data in autonomous driving, we leverage tracking method to augment the data with a few annotated ground truths. Those tracking results are considered as pseudo ground truths and combined with ground truths to fine-tune the model in novel stage. However, the tracking results are biased towards novel classes, which will cause catastrophic forgetting. By introducing LoRA, we solve the forgetting problem and achieve the highest $\miou$ on both base classes and novel classes.


%
%

\bibliographystyle{IEEEtran}
\bibliography{IEEEabrv,myref}

\end{document}